\documentclass[11pt]{article}

\usepackage[final]{acl}

\usepackage{placeins}
% ...
\usepackage{float}
\usepackage[table]{xcolor}
\definecolor{lightblue}{RGB}{220, 235, 250}  
\definecolor{modelgray}{RGB}{240, 240, 240}
\usepackage{booktabs}
\usepackage{multirow}
\usepackage{graphicx}
\usepackage{arydshln}   
\usepackage{times}
\usepackage{latexsym}
\usepackage{enumitem}

\usepackage{amsmath, amssymb}
\usepackage[ruled,vlined]{algorithm2e}

\usepackage[T1]{fontenc}

\usepackage[utf8]{inputenc}
\usepackage{tcolorbox}

\usepackage{microtype}

\usepackage{inconsolata}
\usepackage{graphicx}
\usepackage{graphicx}
\usepackage{hyperref}
\usepackage{tabularx}

\title{Hidden States as Early Signals: Step-level Trace Evaluation and \\ Pruning for Efficient Test-Time Scaling}

\author{Zhixiang Liang\thanks{Equal contribution.}, Beichen Huang\footnotemark[1], Zheng Wang, Minjia Zhang \\
University of Illinois Urbana-Champaign\\
\texttt{\{zliang18, beichen8, zhengw10, minjiaz\}@illinois.edu}}

\usepackage{xspace}
\newcommand{\methodname}{STEP\xspace}
\begin{document}
\maketitle
\begin{abstract}
Large Language Models (LLMs) can enhance reasoning capabilities through test-time scaling by generating multiple traces. However, the combination of lengthy reasoning traces with multiple sampling introduces substantial computation and high end-to-end latency. Prior work on accelerating this process has relied on similarity-based or confidence-based pruning, but these signals do not reliably indicate trace quality. To address these limitations, we propose \textbf{\methodname}: \underline{\textbf{S}}tep-level \underline{\textbf{T}}race \underline{\textbf{E}}valuation and \underline{\textbf{P}}runing, a novel pruning framework that evaluates reasoning steps using hidden states and dynamically prunes unpromising traces during generation. We train a lightweight step scorer to estimate trace quality, and design a GPU memory-aware pruning strategy that triggers pruning as the GPU memory is saturated by KV cache to reduce end-to-end latency. Experiments across challenging reasoning benchmarks demonstrate that \methodname reduces end-to-end inference latency by 45\%–70\% on average compared to self-consistency while also improving reasoning accuracy. 
Our code is released at: \url{https://github.com/Supercomputing-System-AI-Lab/STEP}
\end{abstract}

\section{Introduction}

Large Language Models (LLMs) have demonstrated exceptional reasoning capabilities, particularly through test-time scaling (TTS) techniques that allocate additional computation during inference \cite{wei2022chain,kojima2023largelanguagemodelszeroshot,openai2024gpt4technicalreport,deepseekai2025deepseekr1incentivizingreasoningcapability,hou2025t}. 
Among these methods, self-consistency \cite{wang2022self} is the most widely adopted parallel scaling approach, which generates multiple traces and selects the final answer through majority voting, repeatedly achieving state-of-the-art performance on reasoning tasks \cite{openai2024o1}. However, both lengthy reasoning traces and multiple sampling paths contribute to prohibitive computational costs and substantial latency, severely limiting their practical deployment \cite{wang-etal-2025-faster,ji2025seerselfconsistencyadvancebudget}. Furthermore, self-consistency treats all reasoning trajectories equally, wasting resources on erroneous traces that fail to contribute to the correct answer \cite{hong-etal-2025-slim}.

\begin{figure}[t]
    \centering
    \includegraphics[width=1\linewidth]{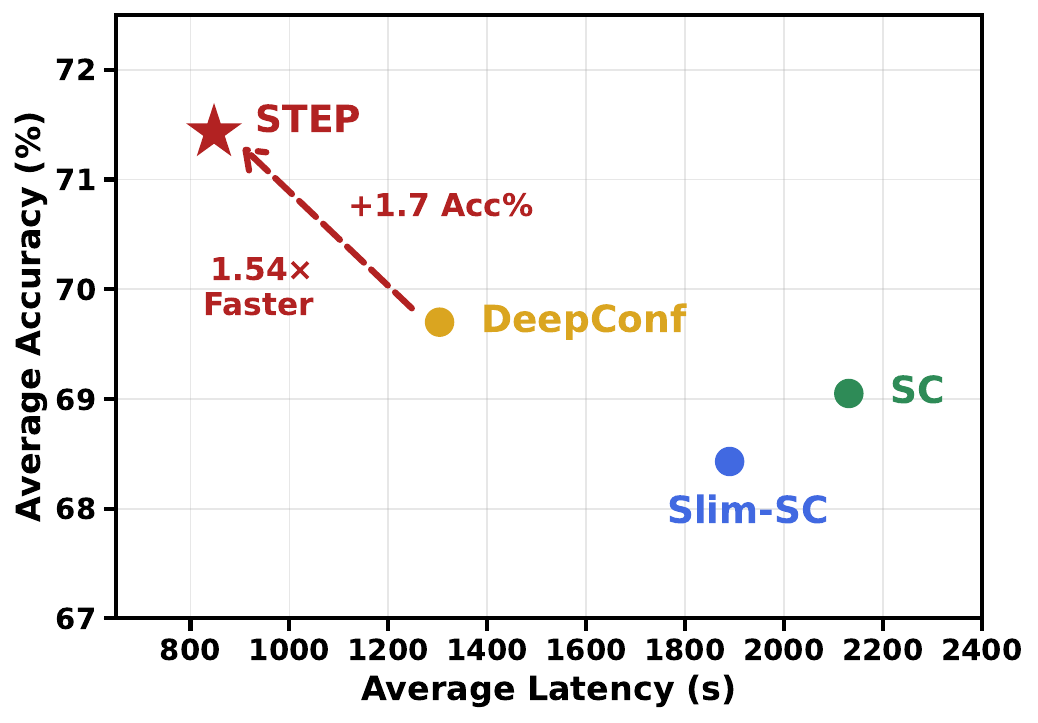}
    \caption{Comparison of accuracy versus latency across different methods on DeepSeek-R1-0528-Qwen3-8B. \methodname achieves superior accuracy (averaged across AIME-25, HMMT-24/25, GPQA-D) while significantly reducing latency compared to baseline methods.}
    \label{fig:figure1}
\end{figure}

Prior work to speed up parallel scaling focuses on pruning low-quality traces during the reasoning process, but faces two fundamental limitations. First, the signals used to evaluate trace quality are  unreliable. One line of methods prunes similar reasoning traces to preserve answer diversity \cite{hong-etal-2025-slim, tu2025deeppruneparallelscalingintertrace}. This is problematic since multiple valid paths can converge to the same correct answer, and surface-level textual similarity does not necessarily indicate redundancy in reasoning quality. Another line of methods leverages the LLM's internal confidence for early stopping \cite{fu2025deepthinkconfidence,kang2025scalablebestofnselectionlarge}, assuming high confidence correlates with correctness. However, confidence scores do not reliably indicate correctness, as models can exhibit high confidence even for factually false or logically inconsistent outputs, which is known as miscalibration \cite{chhikara2025mindconfidencegapoverconfidence}.

Second, existing approaches overlook a critical factor: the dominant bottleneck of end-to-end latency lies not only in the number of generated tokens, but also in the inference system design and its interaction with the algorithm. By focusing primarily on reducing token generation, these methods can yield speedups but fail to fully address the latency bottleneck. When applying parallel scaling methods to complex reasoning tasks, the KV cache of multiple long traces can rapidly exhaust GPU memory. Once the pre-allocated KV cache memory becomes insufficient, inference systems typically preempt traces into a waiting queue until others complete  \cite{kwon2023efficient}. We observe that these waiting periods, together with the resulting redundant computation, constitute the primary end-to-end latency bottleneck.

To address these limitations, we propose \textbf{\methodname}: \underline{\textbf{S}}tep-level \underline{\textbf{T}}race \underline{\textbf{E}}valuation and \underline{\textbf{P}}runing, a novel pruning method that leverages the hidden state to evaluate the trace quality during generation and trigger pruning with GPU memory awareness. Our approach is motivated by two insights from preliminary experiments. First, hidden states at reasoning step boundaries encode rich information about the model's reasoning dynamics \cite{yang2025specexitacceleratinglargereasoning}, making them suitable for quality assessment. Second, these signals emerge early: hidden states from early reasoning steps are already sufficient to distinguish promising traces from unpromising ones. Based on these insights, we train a lightweight step scorer on step boundary hidden states, enabling early assessment of reasoning quality with negligible overhead and allowing us to precisely halt unpromising responses.



Beyond algorithmic design, we additionally consider efficiency from an inference system perspective, which has been largely overlooked by prior work. We leverage GPU memory utilization as the signal to trigger pruning. As KV cache accumulation drives GPU memory toward saturation, we prune the least promising trace and immediately release its resources, preventing preemption and queuing delays. By accounting for this system-level bottleneck, our design reduces unnecessary computation and eliminates memory-induced waiting overhead, substantially improving end-to-end generation latency. 

We evaluate \methodname across challenging reasoning benchmarks (AIME-25, HMMT-24/25, GPQA-Diamond, EquiBench, DivLogicEval) and models at different scales (Qwen3-4B-Thinking-2507, DeepSeek-R1-0528-Qwen3-8B, Phi-4-reasoning-plus(14B)). Experiments demonstrate that \methodname reduces end-to-end inference latency by 45\%–70\% on average compared to self-consistency while also improving reasoning accuracy by +0.4 to +7.5 percentage points. These gains benefit from both the hidden-state-based early pruning and the GPU memory-aware system optimization, highlighting the potential of efficient test-time scaling to advance complex reasoning capabilities in LLMs.





\section{Related Work}

\subsection{Trace Pruning for Parallel Scaling}

Recent work on accelerating parallel scaling has explored pruning unpromising traces during generation. These methods fall into two categories: confidence-based methods that prune low-confidence traces \cite{fu2025deepthinkconfidence,zhou2025bridginginternalprobabilityselfconsistency}, and diversity-based methods that remove similar traces to preserve answer diversity \cite{hong-etal-2025-slim, tu2025deeppruneparallelscalingintertrace}. However, both approaches have notable limitations: confidence scores can suffer from model overconfidence and miscalibration \cite{chhikara2025mindconfidencegapoverconfidence}, while surface-level similarity does not necessarily indicate reasoning redundancy, risking the inadvertent removal of correct traces.
And these primarily aim at reducing the average generated tokens per question to accelerate reasoning process, which neglects the system perspective that fundamentally slows down the generation.


\begin{figure*}[!htbp]
    \centering
    \makebox[\textwidth][c]{%
        \includegraphics[width=1.0\linewidth]{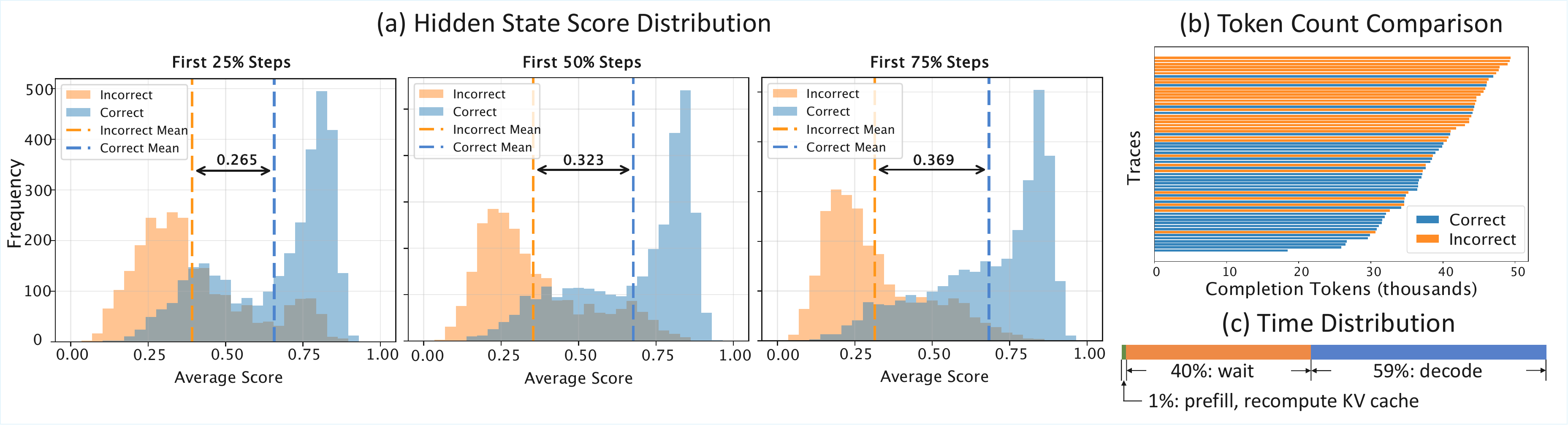}
    }
    \caption{(a) Hidden state score distributions for correct vs. incorrect reasoning traces on HMMT-25. Scores are computed by the scorer model as averages over the first 25\%, 50\%, and 75\% of reasoning steps. (b) Token count comparison of correct and incorrect traces for AIME-25 Q28 using Qwen3-4B-Thinking-2507; incorrect traces average 42.5k tokens compared to 35.3k for correct ones. (c) The time distribution for generating one trace on the same setup; waiting time (40\%) is substantial, with actual decoding occupying 59\%.}
    
    \label{fig:score_distribution}
\end{figure*}

\subsection{Hidden-state as Reasoning Evaluator}
Assessing reasoning trace quality is critical for enhancing the reliability of LLMs in parallel scaling \cite{liang-etal-2025-fine}. Recent studies have investigated using LLM internal representations to assess reasoning quality. \citet{zhang2025reasoning} show that reasoning models encode correctness-related information in their hidden states and that a lightweight probe can predict whether an intermediate answer is correct. CLUE \cite{liang2025cluenonparametricverificationexperience} proposes a non-parametric verifier that clusters hidden-state features from past experience and predicts correctness by comparing a candidate trace's hidden-state signature to success/failure centroids, demonstrating that hidden states provide a strong correctness signal and can improve final selection. 
Building on these findings, we identify step boundaries as natural checkpoints where hidden states provide clear quality signals. We train a lightweight step scorer on these boundary representations, enabling continuous monitoring and early pruning in parallel scaling.


\section{Motivation}




As discussed above, current parallel scaling methods face two key challenges: difficulty in distinguishing correct reasoning traces from incorrect ones and high end-to-end latency. In our preliminary experiments, we find that the hidden states of reasoning models contain rich signals indicative of reasoning quality, and we identify the primary source of the latency bottleneck in parallel scaling.

\paragraph{Discriminative Signals in Hidden States} \label{para:motivation_hs}

Recent studies \cite{zhang2025reasoning, liang2025cluenonparametricverificationexperience} have shown that the hidden states of completed reasoning traces can serve as a reliable proxy for assessing reasoning quality. 
Our preliminary experiments further demonstrate that hidden states from early reasoning steps are already information-rich and provide a strong signal for reasoning correctness.  As illustrated in Fig.~\ref{fig:score_distribution}a, we train a simple 2-layer MLP on hidden states to predict reasoning correctness.
We find that the hidden state score from early steps effectively distinguishes between correct and incorrect reasoning paths, with higher scores indicating a greater likelihood of correctness, and the discriminability becomes stronger as reasoning progresses. These findings indicate that internal signals can efficiently assess reasoning quality during generation, motivating our use of hidden states for early trace pruning.

\paragraph{Latency Bottleneck in Parallel Scaling} 
\label{para:motivation_latency}
We observe two main sources of inefficiency in parallel scaling. First, incorrect traces tend to be longer than correct ones as shown in Fig.~\ref{fig:score_distribution}b. Early termination of such traces can eliminate tokens generated for each question, therefore improving both efficiency and accuracy.
Second, we observe a more fundamental bottleneck arising from inference system behavior, which has been largely overlooked by prior work. As generation progresses, KV cache accumulation quickly saturates GPU memory, causing the inference engine (e.g., vLLM \cite{kwon2023efficient}, SGLang \cite{zheng2024sglang}) to preempt waiting traces by freeing or offloading their KV caches; once an active trace finishes and releases its KV cache, a preempted trace is resumed with its KV cache reconstructed. As shown in Fig.~\ref{fig:score_distribution}c, waiting time accounts for approximately 40\% of end-to-end latency, while actual decoding occupies 59\%. These observations motivate a pruning strategy that explicitly accounts for inference system.
\section{\methodname}


\begin{figure*}[t]
    \centering
    \scalebox{1}[1]{%
        \includegraphics[width=\linewidth]{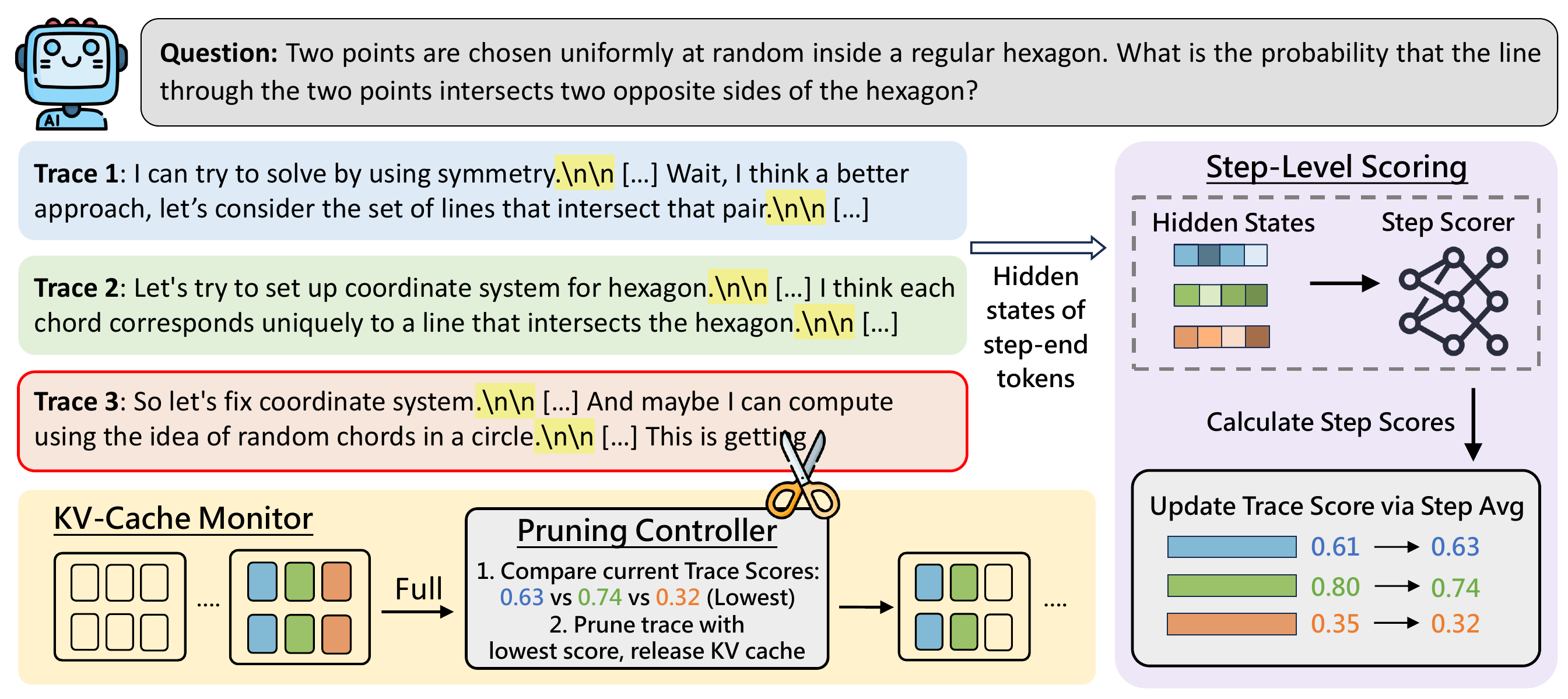}
    }
    \caption{Overview of the \methodname framework. The step-level scoring module extracts hidden states at step boundaries and uses a trained step scorer to compute step-level scores, which are averaged to obtain trace-level scores. The KV-cache monitor triggers pruning when GPU memory is saturated, removing the trace with the lowest score and releasing its KV cache to prevent queuing delays.}
    \label{fig:main_figure}
\end{figure*}

In this section, we progressively construct \methodname. Designing an effective pruning method involves two key questions: \textbf{which reasoning traces to prune} and \textbf{when pruning should be triggered}. As illustrated in the overview in Fig.~\ref{fig:main_figure}, \methodname addresses the first question with a step scorer that evaluates every step during generation, and the second with a KV cache–aware monitoring mechanism. We next describe each component in detail by systematically answering these two questions.





\subsection{Step Scorer}\label{sec:step_scorer}

As discussed in the Section~\ref{para:motivation_hs}, we train a step scorer that leverages hidden-state representations to assess reasoning quality at each step. 


\paragraph{Step Representation}
Following common practice \cite{yang2025speculative}, we extract the reasoning content between “\texttt{\textless think\textgreater}” and “\texttt{\textless/think\textgreater}”, and segment it into $N$ reasoning steps using “\texttt{\textbackslash n\textbackslash n}” as the delimiter. Then a trace is defined as: $t=(s_1,s_2,\ldots,s_N)$. For each step $s_n$, we use the last-layer hidden state of step-end token\footnote{It refers to any token whose text contains "\texttt{\textbackslash n\textbackslash n}". 
}
$\mathbf{h}_n$ 
as input to the scorer, as it accumulates contextual information from all previous reasoning steps in the trace.

\paragraph{Label Construction}
As supervision for the step scorer, we propagate the trace-level correctness label $y \in \{0, 1\}$ to all steps within the trace as pseudo-labels for simplicity, as fine-grained step-level annotation is costly to obtain. Specifically, 
\begin{equation*}
    \tilde{y}_n = y, \quad \forall n \in \{1, \ldots, N\},
\end{equation*}
where $y = 1$ indicates a correct trace and $y = 0$ an incorrect one. For training data curation, we balance the number of correct and incorrect traces, while including all steps from each trace in the training set. Details of the training data are provided in Section~\ref{para:label_construction}.

\paragraph{Model Architecture}
The step scorer $f_\theta$ is a two-layer MLP, which we find sufficient for capturing quality signals from hidden states. It maps $\mathbf{h}_n$ to a correctness probability score $\hat{y}_n$:
\begin{equation*}
\hat{y}_n = f_\theta(\mathbf{h}_n)
= \sigma\!\Big(
\mathbf{W}_2\,\mathrm{ReLU}(\mathbf{W}_1 \mathbf{h}_n + \mathbf{b}_1) + b_2
\Big),
\end{equation*}
where $\mathbf{W}_1$, 
$\mathbf{W}_2$, 
$\mathbf{b}_1 $, 
and $b_2$ are trainable parameters. The function $\sigma(\cdot)$ denotes the sigmoid activation. More details about the scorer model architecture are provided in Appendix~\ref{sec:step_scorer_train}.

\paragraph{Training Objective}
We train the step scorer using a weighted binary cross-entropy loss:
\begin{equation*}
\mathcal{L} =
-\frac{1}{N}\sum_{n=1}^{N}\Big(
\alpha\tilde{y}_n \log \hat{y}_n + (1-\tilde{y}_n)\log(1-\hat{y}_n)
\Big),
\end{equation*}
where $\alpha = K^{-}/K^{+}$ is the ratio of negative to positive samples in the training data. This weighting compensates for the imbalance at the step level, as incorrect traces tend to be longer and thus generate more negative step instances, even when the dataset is balanced at the trace level.

\subsection{Memory Constraint as Trigger} \label{sec:memory_trigger}
The timing of pruning is critical for improving generation efficiency. Prior approaches typically rely on predefined confidence thresholds \cite{fu2025deepthinkconfidence} or fixed wall-clock schedules \cite{hong-etal-2025-slim} to trigger pruning, without considering the behavior of inference system. While these methods reduce generation time by terminating unpromising traces that may produce longer sequences, they overlook a more fundamental bottleneck revealed in Section \ref{para:motivation_latency}: the excessive waiting time caused by GPU memory constraints. As a result, existing methods fail to address the dominant source of inefficiency during inference.

To overcome this limitation, we propose a GPU memory-triggered pruning mechanism. Whenever GPU memory is full and the KV cache for the next decoding step cannot be scheduled, we immediately prune the trace with the lowest average step score and release its KV cache. This design completely eliminates waiting queues, thereby avoiding prolonged suspension and repeated resumption of traces. The greedy strategy here is simple to implement and easy to interpret, while leading to strong empirical improvements.
\begin{table*}[htbp]
\renewcommand{\arraystretch}{1.2}
\centering
\setlength{\tabcolsep}{3.2pt}
\small
\begin{tabular}{|l|ccc|ccc|ccc|ccc|ccc|}
\hline
\multirow{2}{*}{\textbf{Methods}} & \multicolumn{3}{c|}{\textbf{AIME-25}} & \multicolumn{3}{c|}{\textbf{HMMT-24/25}} & \multicolumn{3}{c|}{\textbf{GPQA-D}} & \multicolumn{3}{c|}{\textbf{EquiBench}} & \multicolumn{3}{c|}{\textbf{DivLogicEval}} \\
& \textit{Acc.}$\uparrow$ & \textit{Tok.}$\downarrow$ & \textit{Lat.}$\downarrow$ & \textit{Acc.}$\uparrow$ & \textit{Tok.}$\downarrow$ & \textit{Lat.}$\downarrow$ & \textit{Acc.}$\uparrow$ & \textit{Tok.}$\downarrow$ & \textit{Lat.}$\downarrow$ & \textit{Acc.}$\uparrow$ & \textit{Tok.}$\downarrow$ & \textit{Lat.}$\downarrow$ & \textit{Acc.}$\uparrow$ & \textit{Tok.}$\downarrow$ & \textit{Lat.}$\downarrow$ \\
\hline
\rowcolor{modelgray} \multicolumn{16}{|c|}{\textit{Qwen3-4B-Thinking-2507}} \\
\hdashline
CoT & 81.3 & 22.7 & 145 & 51.7 & 28.3 & 184 & 65.8 & 8.9 & 54 & 67.2 & 7.8& 41& 51.0& 8.7 & 49 \\
SC & 86.7 & 1454.3 & 1430 & 57.9 & 1809.9 & 2055 & 68.1 & 569.1 & 252 & 70.4& 498.9 & 237 & 54.3 & 554.7& 228\\
Slim-SC & 86.7 & 957.5 & 767 & 57.9 & \textbf{966.7} & 937 & 64.9 & 414.7 & 236 & 73.7& 445.8& 232 & 54.8& 547.6 & 240\\
DeepConf & \textbf{90.0} & \textbf{841.5} & 933 & 62.5 & 1053.2 & 1313 & 67.6 & \textbf{379.1} & 257 & 71.5 & \textbf{379.5}& 324 & 53.8 & \textbf{313.8}& 296 \\
\rowcolor{lightblue} \methodname & 88.3 & 1131.5 & \textbf{675} & \textbf{64.2} & 1129.6 & \textbf{856} & \textbf{68.5} & 539.6 & \textbf{223} & \textbf{74.0}& 432.1& \textbf{214} &  \textbf{55.7}& 509.3& \textbf{209}\\
\hline
\rowcolor{modelgray} \multicolumn{16}{|c|}{\textit{DeepSeek-R1-0528-Qwen3-8B}} \\
\hdashline
CoT & 77.5 & 26.4 & 204 & 55.2 & 31.5 & 282 & 62.3 & 11.4 & 81 & 69.5& 5.3 &40 &39.0 & 5.7 &44 \\
SC & 83.3 & 1691.0 & 2259 & 62.9 & 2014.6 & 2891 & 67.1 & 729.8 & 484 & 75.6& 331.5 &189 &44.1 & 363.5 &192 \\
Slim-SC & 83.3 & 1519.9 & 1960 & 62.1 & 1782.0 & 2589 & 66.2 & 564.1 & 424 & 75.0 & 341.3& 177 &45.0 &  361.8&180 \\
DeepConf & 81.7 & \textbf{916.4} & 1475 & 64.2 & \textbf{1038.7} & 1666 & \textbf{68.7} & \textbf{419.8} & 409 & 74.8 & \textbf{232.2}  &221 &45.2 & \textbf{276.4} &202 \\
\rowcolor{lightblue} \methodname & \textbf{85.0} & 989.7 & \textbf{891} & \textbf{66.3} & 1096.5 & \textbf{1061} & 68.2 & 635.7 & \textbf{378} & \textbf{77.3}& 282.8 & \textbf{173}&\textbf{45.6} &293.7  &\textbf{162 }\\
\hline
\rowcolor{modelgray} \multicolumn{16}{|c|}{\textit{Phi-4-reasoning-plus}} \\
\hdashline
CoT & 78.3 & 16.0 & 194 & 55.2 & 21.5 & 270 & 69.5 & 11.9 & 105 &62.0 & 12.1& 108& 42.3 & 8.2& 98 \\
SC & 86.7 & 1026.7 & 1687 & 65.9 & 1373.1 & 2467 & 76.3 & 762.5 & 1081 & 66.2& 772.3 & 929& \text{46.7}& 520.4 & 445 \\
Slim-SC & 85.0 & 875.8 & 1354 & 64.6 & 1149.7 & 1804 & 72.3 & 560.6 & 655 & 65.8 & 578.4 & 603 & 45.3& 463.6& 433\\
DeepConf & 85.8 & 537.2 & 1165 & 66.3 & 735.3 & 1647 & 74.8 & \textbf{401.9} & 1285 & 64.5 & \textbf{396.0} & 718 & 45.8 & \textbf{284.7} & 402 \\
\rowcolor{lightblue} \methodname & \textbf{87.5} & \textbf{503.4} & \textbf{519} & \textbf{67.1} & \textbf{582.5} & \textbf{637} & \textbf{76.7} & 441.5 & \textbf{445} & \textbf{67.9}& 453.8 & \textbf{421} & \textbf{47.0} & 423.2&  \textbf{319}\\
\hline
\end{tabular}
\caption{Main experimental results comparing our method with baseline methods (CoT, SC, Slim-SC, and DeepConf) on various models and benchmarks. Evaluation metrics include accuracy (\%), average token usage ($\times 10^{3}$), and inference latency (s). The HMMT-24/25 column reports results averaged across the HMMT-24 and HMMT-25 benchmarks.}
\label{tab:results}
\end{table*}

\subsection{Pruning with \methodname}

With the step scorer (Section~\ref{sec:step_scorer}) and memory-aware pruning trigger (Section~\ref{sec:memory_trigger}) in place, we now describe the overall algorithm, whose pseudo-code is shown in  Algorithm \ref{alg:online_kv_prune}.

\begin{algorithm}[h]
\small
\caption{\methodname}
\label{alg:online_kv_prune}
\KwIn{Problem $P$, step scorer $f(\cdot)$, trace budget $N$}
\KwOut{Final answer $\hat{a}$}

$\mathcal{T} \leftarrow \textsc{InitTraces}(P, N)$

\While{$\mathcal{T} \neq \emptyset$}{
    \ForEach{trace $t \in \mathcal{T}$}{
        $(x, h) \leftarrow \textsc{NextToken}( t)$\;
        \If{"\texttt{\textbackslash n\textbackslash n}" in $x$}{
            $\hat{y} \leftarrow f(h)$\;
            Update $score_t$ with $\hat{y}$
        }

        \If{\textsc{GPUMemoryFull}}{
            $t^\star \leftarrow \arg\min_{t \in \mathcal{T}} \; score_t$\;
            \textsc{ReleaseKVCache}$(t^\star)$\;
            $\mathcal{T} \leftarrow \mathcal{T} \setminus \{t^\star\}$\;
        }
    }
        
}

$\hat{a} \leftarrow \textsc{WeightedVote}(\mathcal{T}, \{score_t\})$\;
\Return{$\hat{a}$}\;
\end{algorithm}

We next detail the score aggregation and answer selection procedures. Given an input prompt, we generate $T$ reasoning traces in parallel. During generation, the step scorer evaluates each reasoning step as it is produced. We aggregate step-level scores into a trace-level score via a running average:

$$\text{score}_t = \frac{1}{n}\sum_{i=1}^{n} \hat{y}_i^t$$

where $n$ is the number of steps generated so far in trace $t$. We adopt the average score rather than the latest step score alone, as it captures the evolution of reasoning quality across steps and is less sensitive to individual step variance. Once all traces have either completed or been pruned, we perform weighted majority voting using the final trace-level scores to determine the answer.

\section{Experiment}

In this section, we conduct comprehensive experiments to evaluate the effectiveness of our method, demonstrating improvements in both reasoning quality and generation efficiency. We further analyze its performance under different settings and investigate the underlying factors contributing to these improvements.
\subsection{Experiment Setup}

\paragraph{Models}

We evaluate on three reasoning LLMs: Qwen3-4B-Thinking-2507 \cite{yang2025qwen3technicalreport}, DeepSeek-R1-0528-Qwen3-8B \cite{deepseekai2025deepseekr1incentivizingreasoningcapability}, and Phi-4-reasoning-plus(14B) \cite{abdin2025phi4reasoningtechnicalreport}. These models are selected for their strong mathematical reasoning and long-chain-of-thought capabilities, and are fully open-sourced to ensure reproducibility.

\paragraph{Benchmarks}

We evaluate on six challenging datasets: AIME-25 \cite{aops2025aime}, HMMT-24 \cite{hmmt_feb_2024_archive}, HMMT-25 \cite{hmmt_feb_2025_archive}, GPQA-Diamond \cite{rein2024gpqa}, EquiBench \cite{wei-etal-2025-equibench}, and DivLogicEval \cite{chung-etal-2025-divlogiceval}. The first three comprise high-difficulty mathematical competition problems, GPQA consists of graduate-level reasoning tasks in general science, EquiBench evaluates code understanding through program equivalence checking, and DivLogicEval benchmarks classical logical reasoning with multiple-choice problems.

\paragraph{Baselines}
We compare our method against the following baseline methods:
\begin{itemize}[itemsep=4pt, topsep=4pt, parsep=0pt, partopsep=0pt]
    \item \textbf{Chain-of-Thought (CoT)} \cite{wei2022chain} uses standard CoT prompting, where the model generates a single reasoning trajectory that directly leads to the final answer.
    \item \textbf{Self-Consistency (SC)} \cite{wang2022self} generates N independent reasoning traces and determines the final answer by majority voting on the predicted solutions. 
        \item \textbf{Slim-SC} \cite{hong-etal-2025-slim} proposes a step-wise thought pruning strategy for self-consistency: it detects and removes redundant reasoning chains by measuring inter-chain similarity at the thought level, reducing latency while maintaining accuracy.
    \item \textbf{DeepConf} \cite{fu2025deepthinkconfidence} utilizes the model’s internal confidence signals to monitor the quality of each reasoning trace during generation, allowing dynamic termination of unpromising traces. The final answer is decided by confidence-weighted voting.
\end{itemize}

\paragraph{Implementation Details} \label{para:label_construction}
To train the step scorer, we curated a dataset of mathematical problems from HMMT 2012--2023 \cite{hmmt2012_2023}, which provide diverse examples for learning hidden state representations indicative of reasoning quality. We sampled 64 solutions from the target model for each problem and verified their correctness using a deterministic rule-based verifier. We then randomly selected 5,000 correct and 5,000 incorrect traces to form a balanced training set. More details are shown in Appendix~\ref{sec:step_scorer_train}.

We evaluate all methods under the sampling budget of $N = 64$. For Slim-SC, we apply random pruning with a similarity threshold of 0.95 as recommended in the original work~\cite{hong-etal-2025-slim}. For DeepConf, we use the online variant (DeepConf-low) with $N_{\text{init}} = 16$ traces for offline warmup, then generate the remaining 48 traces with early termination for those falling below the top-10\% confidence threshold.
All experiments are conducted using a modified vLLM~\cite{kwon2023efficient} framework with our pruning algorithm on a single 96GB NVIDIA GH200 GPU. 
More detailed settings are provided in the Appendix~\ref{sec:appendix_imple_detail}.


\begin{figure*}[t]
    \centering
    \includegraphics[width=1\linewidth]{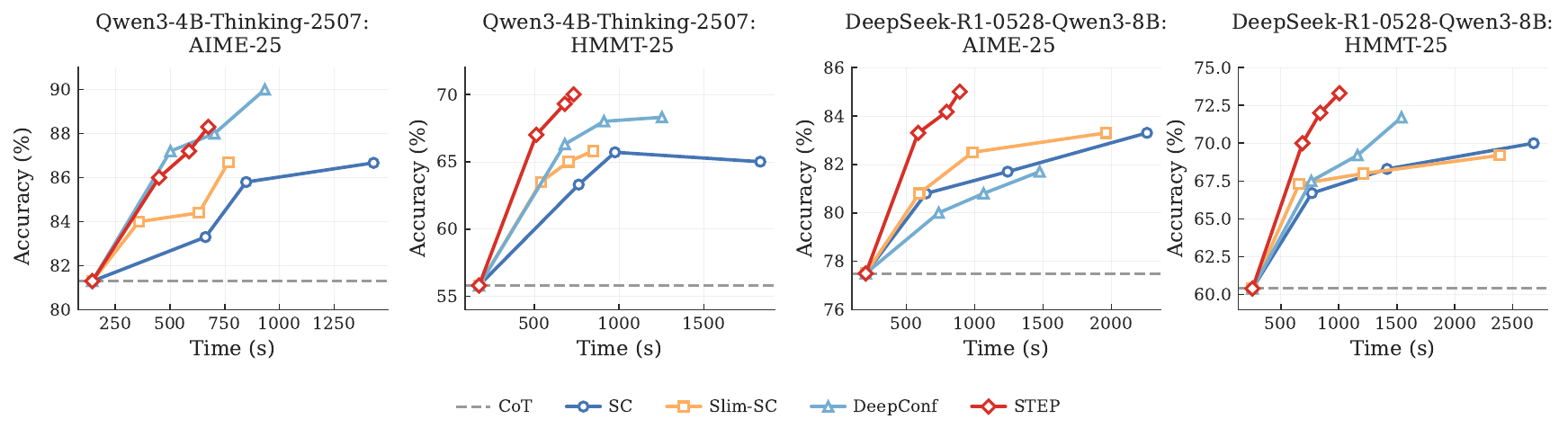}
    \caption{Latency scaling comparison between \methodname and baseline methods on AIME-25 and HMMT-25 using Qwen3-4B-Thinking-2507 and DeepSeek-R1-0528-Qwen3-8B. The data points here represent the accuracy and latency for a sampling budget of N = 1,16,32, and 64, ordered from left to right.}
    \label{fig:latency_scaling}
\end{figure*}
\subsection{Main Results}
Tab.~\ref{tab:results} presents the main experimental results comparing our method against baseline approaches across six benchmarks and three reasoning models. We report accuracy, average output token usage, and latency per problem as evaluation metrics.

\paragraph{Consistent Accuracy Improvements} Our method achieves the highest accuracy on most of the benchmark-model combinations. On mathematical reasoning benchmarks, our approach consistently outperforms SC, Slim-SC, and DeepConf across all three models. For example, on HMMT-25, our method improves accuracy by 5.0\%, 3.3\%, and 0.8\% over SC for Qwen3-4B-Thinking-2507, DeepSeek-R1-0528-Qwen3-8B, and Phi-4-reasoning-plus, respectively. For the general science reasoning benchmark GPQA-Diamond, our method also achieves competitive accuracy across all models, demonstrating its generalizability beyond mathematical domains. We attribute \methodname's accuracy improvements to our hidden-state-based step scorer, which provides accurate estimation of trace quality during generation, enabling effective pruning of unpromising traces and reliable answer aggregation through weighted voting. We provide a detailed analysis of our step scorer's ranking ability in Section~\ref{sec:rank_ability}.

\paragraph{Superior Computational Efficiency} A key advantage of our method lies in end-to-end latency improvement. Compared to SC, our approach reduces latency by 45\%–70\% on average across different settings. For instance, on Phi-4-reasoning-plus with HMMT-24, our method achieves 58.3\% accuracy in just 630 seconds, compared to 2405 seconds for SC, yielding a 3.8× speedup. Even compared to Slim-SC and DeepConf, our method consistently achieves lower latency while maintaining comparable or superior accuracy. On DeepSeek-R1-0528-Qwen3-8B with AIME-25, our approach reduces latency from 1475s (DeepConf) to 891s, a 1.7× speedup, while achieving higher accuracy. And from result of average output token length, we observe that our method consistently reduces the token counts compared with SC across all models and benchmarks, while keeping a comparable level of tokens with DeepConf and Slim-SC. These results reveal that eliminating waiting queue significantly contributes to accelerating generation. We present a detailed analysis of acceleration at Section \ref{sec:profile_acc}.

\subsection{Analysis}

\subsubsection{Latency Scaling}

To verify the effectiveness and efficiency of \methodname under varying computational budgets, we conduct latency scaling experiments using Qwen3-4B-Thinking-2507 and DeepSeek-R1-0528-Qwen3-8B on AIME-25 and HMMT-25. We set the sampling budget to 16, 32, and 64, and evaluate our method against baseline approaches. Since larger budgets increase both latency and potential accuracy, this setup allows us to examine accuracy-latency trade-offs across different computational regimes. 

As illustrated in Fig.~\ref{fig:latency_scaling}, \methodname  achieves superior accuracy-latency trade-offs in most settings, reaching higher accuracy at any given time budget. For instance, on Qwen3-4B-Thinking-2507 with HMMT-25, \methodname achieves 70\% accuracy using around 40\% of the latency required by SC, which reaches just 65\% accuracy. Similarly, on DeepSeek-R1-0528-Qwen3-8B with AIME-25, our method attains 85\% accuracy while consuming approximately 40\% of the latency that SC requires to achieve comparable performance. Compared to Slim-SC and DeepConf, which also aim to reduce inference cost, our method still demonstrates clear advantages. On HMMT-25, our method consistently outperforms both Slim-SC and DeepConf across all time budgets, achieving higher accuracy with lower latency. These results demonstrate that our method exhibits favorable latency scaling properties, outperforming the baseline methods across different models and datasets.

\subsubsection{Ranking Ability of Step Scorer}
\label{sec:rank_ability}

Beyond overall efficiency, we examine whether our step scorer can reliably identify promising traces during the reasoning process. We evaluate its discriminative ability against token-level confidence, as used in DeepConf. We conduct experiments on 256 reasoning traces per question generated by Qwen3-4B-Thinking-2507 on AIME-25 and HMMT-25. For each trace, we input only the first $k\%$ of steps and compute an average score. We compare against mean token-level confidence computed from the same partial trace. We adopt pairwise ranking accuracy (RankAcc) as our evaluation metric. For each question $q$ with a set of correct traces $\mathcal{P}_q$ and incorrect traces $\mathcal{N}_q$, RankAcc measures the proportion of correctly ordered positive--negative pairs:
\begin{equation*}
\text{RankAcc} = \mathbb{E}_{q \in Q} \left[ \mathbb{E}_{p \in \mathcal{P}_q, n \in \mathcal{N}_q} \left[ \mathbb{1}[s(p) > s(n)] \right] \right],
\end{equation*}
where $s(\cdot)$ denotes the scoring function.

\begin{figure} [t]
    \centering
    \includegraphics[width=1.0\linewidth]{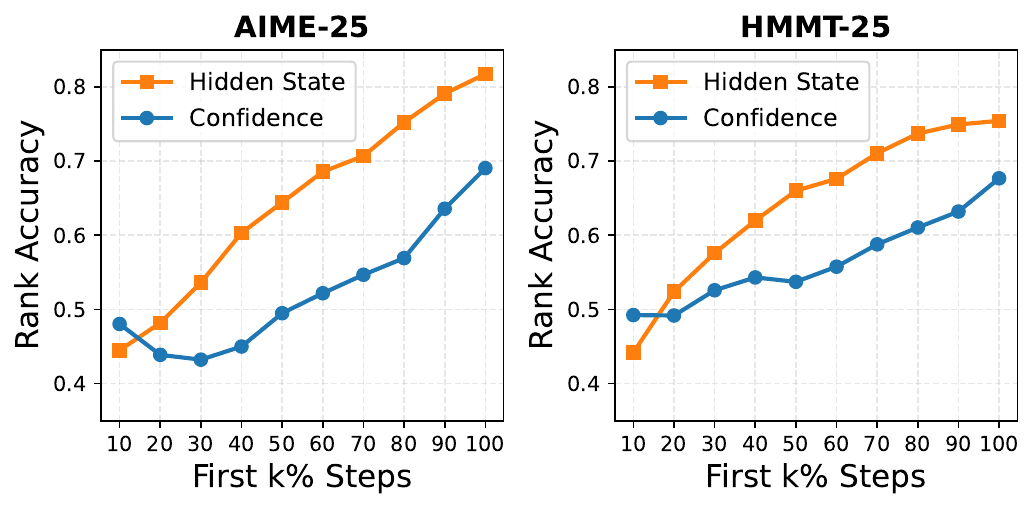}
    \caption{Pairwise RankAcc of the hidden-state-based step scorer versus token-level confidence. 
    }
    \label{fig:rankacc}
\end{figure}

As shown in Fig.~\ref{fig:rankacc}, the hidden-state-based step scorer outperforms token confidence, and its discriminative performance improves steadily as more reasoning steps become available. Even at early stages, the step scorer already achieves strong ranking accuracy, demonstrating that hidden states encode rich information about solution correctness well before the final answer is reached. We also include a visualization of trace-level score dynamics from hidden-state-based step scorer in Appendix~\ref{sec:score_dynamics}.


\subsubsection{Comparison with Reward Models}

To further validate the effectiveness of our step scorer, we compare it against Qwen2.5-Math-PRM-7B \cite{zhang-etal-2025-lessons}, one of the strongest open-source process reward models. Using the same set of 64 reasoning traces across 4 independent runs, we compare three answer aggregation strategies: majority voting, PRM-weighted voting, and our STEP scorer-weighted voting.

\begin{table}[htbp]
\centering
\small
\resizebox{0.48\textwidth}{!}{
\begin{tabular}{l ccc}
\toprule
\textbf{Voting Method} & \textbf{AIME-25} & \textbf{HMMT-25} & \textbf{GPQA-D} \\
\midrule
\multicolumn{4}{l}{\textit{Qwen3-4B-Thinking-2507}} \\[3pt]
Majority Voting        & 86.7 & 65.0 & 68.1 \\
PRM Weighted & 87.5   & 67.5   & 68.7   \\
STEP Weighted     & \textbf{90.0}   & \textbf{71.7}   & \textbf{69.2}   \\
\midrule
\multicolumn{4}{l}{\textit{DeepSeek-R1-0528-Qwen3-8B}} \\[3pt]
Majority Voting      & 83.3 & 70.0 & 67.1 \\
PRM Weighted & 83.3   & 71.7   &  66.4  \\
STEP Weighted     & \textbf{85.0}   & \textbf{75.8}   &\textbf{68.5}   \\
\bottomrule
\end{tabular}
}
\caption{Accuracy (\%) comparison of different voting strategies on the same set of reasoning traces.}
\label{tab:prm_comparison}
\end{table}

As shown in Tab.~\ref{tab:prm_comparison}, STEP-weighted voting consistently outperforms both baselines across all benchmark-model combinations, surpassing PRM-weighted voting by up to 4.2\% on individual benchmarks. Notably, our step scorer is a simple 2-layer MLP operating on hidden states that are naturally produced during generation, whereas the PRM requires a separate 7B-parameter model with a full forward pass over each trace. The scorer is trained on only 10,000 automatically verified traces and introduces negligible overhead ($<10^{-6}$ relative FLOPs per step as shown in Appendix~\ref{sec:appendix_additional_computational_overhead}), demonstrating that hidden states provide an efficient and effective alternative to external reward models.




\subsubsection{Profiling Acceleration by Pruning}
\label{sec:profile_acc}
The acceleration of \methodname stems from two complementary factors: reducing the number of generated tokens and eliminating waiting time during generation. Token counts are reported in Tab.~\ref{tab:results}. We further conduct experiments, profiling the end-to-end generation time breakdown of DeepSeek-R1-0528-Qwen3-8B on HMMT-25 with 64 traces, and results are reported in Tab.~\ref{tab:waiting_decode_time}. 
DeepConf consists of two consecutive stages with N=16 in warmup and N=48 in pruning stage, and we report them separately. We observe that all pruning methods decrease generated tokens compared with SC, and lead to lower decoding time in Tab.~\ref{tab:waiting_decode_time}. Beyond token-level efficiency, the key distinction lies in how to handle waiting time. DeepConf and Slim-SC reduce waiting time compared to SC, since pruning naturally alleviates GPU memory pressure, thus shortening the waiting queue. However, their pruning decisions are not explicitly tied to GPU memory usage, and therefore cannot fully eliminate the waiting time. In contrast, our method completely removes waiting queue by triggering pruning in a memory-aware manner, resulting in the lowest end-to-end generation latency.


\begin{table}[htbp]
\centering
\small
\resizebox{0.48\textwidth}{!}{
\begin{tabular}{lcccccc}
\toprule
  \textbf{Method} &\multirow{2}{*}{\textbf{SC}}  & \multicolumn{2}{c}{\textbf{DeepConf}} & \multirow{2}{*}{\textbf{Slim-SC}}  & \multirow{2}{*}{\textbf{\methodname}}  \\
\cmidrule(lr){1-1} \cmidrule(lr){3-4}
\textbf{Stage} &  & \textbf{Warmup} & \textbf{Prune} &  &  \\
\midrule
\textbf{Wait}   & 1526 & 69  & 194 & 1155 & \text{0} \\
\textbf{Decode} & 1256 & 680 & 726 & 983  & 1024 \\
\bottomrule
\end{tabular}
}
\caption{Waiting time and decoding time (in seconds) comparison across different methods.}
\label{tab:waiting_decode_time}
\end{table}

Taken together, these results demonstrate that while reducing generated tokens lowers decoding overhead, explicitly eliminating waiting time is critical for further accelerating end-to-end generation.

\subsubsection{GPU memory sensitivity}

\methodname triggers pruning when KV cache saturates GPU memory. Since smaller GPU memory budgets lead to earlier pruning, it is important to examine the robustness of our method under different memory constraints. To this end, we conduct a sensitivity analysis by varying the maximum GPU memory utilization from 0.5 to 0.9, on a 96GB NVIDIA GH200 GPU. We report results on the HMMT-25 using DeepSeek-R1-0528-Qwen3-8B, sampling 32 reasoning traces per problem. The results are summarized in Tab.~\ref{tab:gpu_memory_sensitivity}. 

\begin{table}[h]
\centering
\begin{tabular}{lccccc}
\toprule
\textbf{Memory} & 0.5 & 0.6 & 0.7 & 0.8 & 0.9 \\
\midrule
\textbf{Accuracy} & 70.0 & 69.1 & 70.0 & 68.3 & 73.3 \\
\bottomrule
\end{tabular}
\caption{Accuracy result under different GPU memory utilization settings.}
\label{tab:gpu_memory_sensitivity}
\end{table}

We observe that the accuracy remains stable across different memory budgets (70.1 ± 1.8\%). Even under smaller GPU memory budgets, where pruning is triggered earlier, our method consistently achieves strong performance. This observation is consistent with the findings in Section~\ref{sec:rank_ability}, which show that our scorer is able to identify promising reasoning traces at an early stage of generation. These results suggest that our method is insensitive to GPU memory.






\section{Conclusion}
In this work, we introduce \methodname, a method that combines hidden-state-based trace evaluation with GPU memory-aware pruning for efficient test-time scaling. \methodname addresses two fundamental challenges in parallel scaling: unreliable trace quality signals and high end-to-end latency caused by GPU memory saturation. By training a lightweight step scorer on hidden-state representations at reasoning step boundaries, \methodname enables early and accurate identification of unpromising traces during generation. By triggering pruning based on GPU memory utilization, \methodname completely eliminates the waiting queue that dominates end-to-end latency in existing approaches. Experiments across six benchmarks spanning mathematical reasoning, general science, code understanding, and logic reasoning, and three models at different scales, demonstrate that \methodname reduces end-to-end inference latency by 45\%-70\% on average compared to self-consistency while also improving reasoning accuracy by +0.4 to +7.5 percentage points. These results highlight the potential of combining algorithm-level and system-level optimization for efficient parallel scaling in complex reasoning tasks.
\section*{Limitations}
Our work has three primary limitations. First, the step scorer relies on pseudo-labels generated by propagating trace-level correctness to individual steps. This weak supervision assumes that all steps in a correct trace are equally good and all steps in an incorrect trace are equally bad, which is inherently noisy and may not reflect the true quality of individual reasoning steps. Incorporating fine-grained step-level supervision, such as process reward signals or automated step-level verification, could improve the scorer's precision and is a promising direction for future work. Second, the most significant latency improvements depend on memory-triggered pruning, which is tightly coupled to the serving infrastructure. While we observe consistent speedups across different memory budgets, the exact magnitude may differ across inference engines, multi-GPU configurations, and alternative memory management strategies. Third, our current evaluation focuses on tasks with fixed-form verifiable answers. Extending STEP to open-ended generation tasks, such as code generation or long-form writing, would require adapting the answer aggregation mechanism beyond majority voting.
\section*{Acknowledgments}
This research was supported by the National Science Foundation (NSF) under Grant No. 2441601. The work utilized the Delta and DeltaAI system at the National Center for Supercomputing Applications (NCSA) and Jetstream2 at Indiana University through allocation CIS240055 from the Advanced Cyberinfrastructure Coordination Ecosystem: Services \& Support (ACCESS) program, which is supported by National Science Foundation grants \#2138259, \#2138286, \#2138307, \#2137603, and \#2138296. The Delta advanced computing resource is a collaborative effort between the University of Illinois Urbana-Champaign and NCSA, supported by the NSF (award OAC 2005572) and the State of Illinois. UIUC SSAIL Lab is supported by research funding and gift from Google, IBM, Amazon, and AMD, including the Google ML and Systems Junior Faculty Award.


\bibliography{custom}

\appendix

\label{sec:appendix}

\section{Step Scorer Training}
\label{sec:step_scorer_train}
\subsection{Training Parameters}

The training hyper-parameters used in the training process of step scorer are listed in Tab.~\ref{tab:training_params}.

\begin{table}[htbp]
\small
\centering
\begin{tabular}{l|c}
\hline
\textbf{Parameter} & \textbf{Value} \\
\hline
MLP Structure & Input $\rightarrow$ 512 (ReLU) $\rightarrow$ 1 \\
Batch Size & 128 \\
Max Epochs & 20 \\
Early Stopping Patience & 5 \\
Learning Rate & $1 \times 10^{-4}$ \\
Weight Decay & $1 \times 10^{-5}$ \\
Optimizer & Adam \\
Loss Function & BCEWithLogitsLoss \\
\hline
\end{tabular}
\caption{Training Parameters}
\label{tab:training_params}
\end{table}

The input dimension corresponds to the hidden state size of each LLM: 2560 (Qwen3-4B-Thinking-2507), 4096 (DeepSeek-R1-0528-Qwen3-8B), and 5120 (Phi-4-reasoning-plus).

\subsection{Training Dataset}
For training the step scorer, we constructed a dataset comprising mathematical problems from HMMT 2012–2023 \cite{hmmt2012_2023}. We specifically utilized problems from the February competition in Algebra, Combinatorics, and Geometry, which provide diverse and challenging examples for learning hidden state representations indicative of reasoning quality. 

We sampled 64 solutions from the corresponding LLM for each problem and verified the correctness of their final answers using a deterministic rule-based verifier adapted from the Qwen2.5-Math project \cite{yang2024qwen2}. The verifier normalizes answer strings and checks correctness against the ground truth via numeric matching and SymPy-based symbolic equivalence. We then randomly selected 5{,}000 correct and 5{,}000 incorrect traces to form a balanced training set for each LLM.

\section{Experimental Settings}
\label{sec:appendix_imple_detail}

\subsection{Sampling Parameters}

The sampling parameters used for each model across all experiments are listed in Tab.~\ref{tab:sampling_param}. Temperature, top-p, top-k, and maximum generation length remain fixed for all methods. Here, Qwen3-4B refers to Qwen3-4B-Thinking-2507 and Deepseek-8B refers to DeepSeek-R1-0528-Qwen3-8B.

\begin{table}[htbp]
\centering
\resizebox{\columnwidth}{!}{%
\begin{tabular}{lcccc}
\toprule
Model & Temperature & Top-$p$ & Top-$k$ & Max gen len \\
\midrule
Qwen3-4B  & 0.6 & 0.95 & 20 & 64k  \\
DeepSeek-8B      & 0.6 & 0.95 & 20 & 64k  \\
Phi-4-reasoning-plus     & 0.8 & 0.95 & 50 & 32k  \\
\bottomrule
\end{tabular}%
}
\caption{Sampling Parameters}
\label{tab:sampling_param}
\end{table}

\subsection{Prompt Templates}

We apply the following prompt template to each problem in the test benchmarks for all methods.
\begin{tcolorbox}[
    title=Prompt Template,
    colback=white,
    colframe=black,
    coltitle=white,
    colbacktitle=black,
    fonttitle=\bfseries,
    rounded corners,
    boxrule=1pt,
    arc=3mm
]
Please reason step by step, and put your final answer within \verb|\\boxed{}|. \\
Question: \{question\_text\}
\end{tcolorbox}

\subsection{Baselines}
\begin{itemize}[itemsep = 1pt, topsep=5pt]
\item \textbf{Slim-SC}: We use the Random Pruning (RP) strategy in the original paper since it provides a clear advantage in inference speed while retaining accuracy.
\item \textbf{DeepConf}: We use the online variant (DeepConf-low), where traces are terminated when their confidence falls below the level that retains the top 10\% highest-confidence traces from the warmup phase. We set $N_{\text{init}} = 16$ warmup traces for $N \in \text{\{}{32, 64} \text{\}}$ and $N_{\text{init}} = 8$ for $N = 16$.
\end{itemize}

\section{System Integration}
\methodname is built upon the implementation logic of vLLM-V1, where the engine core and model runner reside in separate processes. We place the step scorer on the same GPU and execute its running logic within the same process as the model runner. Only the scores are passed via inter-process communication to the engine core, where the scheduler performs pruning decisions.
The experimental environment is configured as follows:
\begin{itemize}[itemsep = 1pt, topsep=5pt]
\item \textbf{vLLM}: 0.11.1
\item \textbf{Python}: 3.12
\item \textbf{CUDA}: 12.4
\end{itemize}

\section{Additional Computational Overhead}
\label{sec:appendix_additional_computational_overhead}
The step scorer is implemented as an auxiliary MLP, which inevitably introduces additional computation. Since this MLP is invoked at every reasoning step, we quantify its overhead by comparing the per-step computation cost of the scorer with that of the underlying LLM.

We approximate the FLOPs of one forward generation step of the LLM as $2N$, where $N$ denotes the number of non-embedding parameters \cite{kaplan2020scaling}. The computation cost of the step-level MLP is $2m(d+1)$, where $m$ is the hidden dimension of the MLP and $d$ is the hidden dimension of the LLM. The relative overhead per step is therefore
\begin{equation*}
\frac{2m(d+1)}{2N*t},
\end{equation*}
where $t$ is the average tokens per step. In practice, we set $m=512$, $d$ is on the order of $10^3$, $N$ is on the order of billions, and $t$ is around $10^2$. Under these settings, the resulting ratio is below $10^{-6}$, indicating that the computational overhead introduced by the step scorer is negligible.

\section{Trace-level Score Dynamics}
\label{sec:score_dynamics}

\begin{figure*}
    \centering
    \includegraphics[width=1\linewidth]{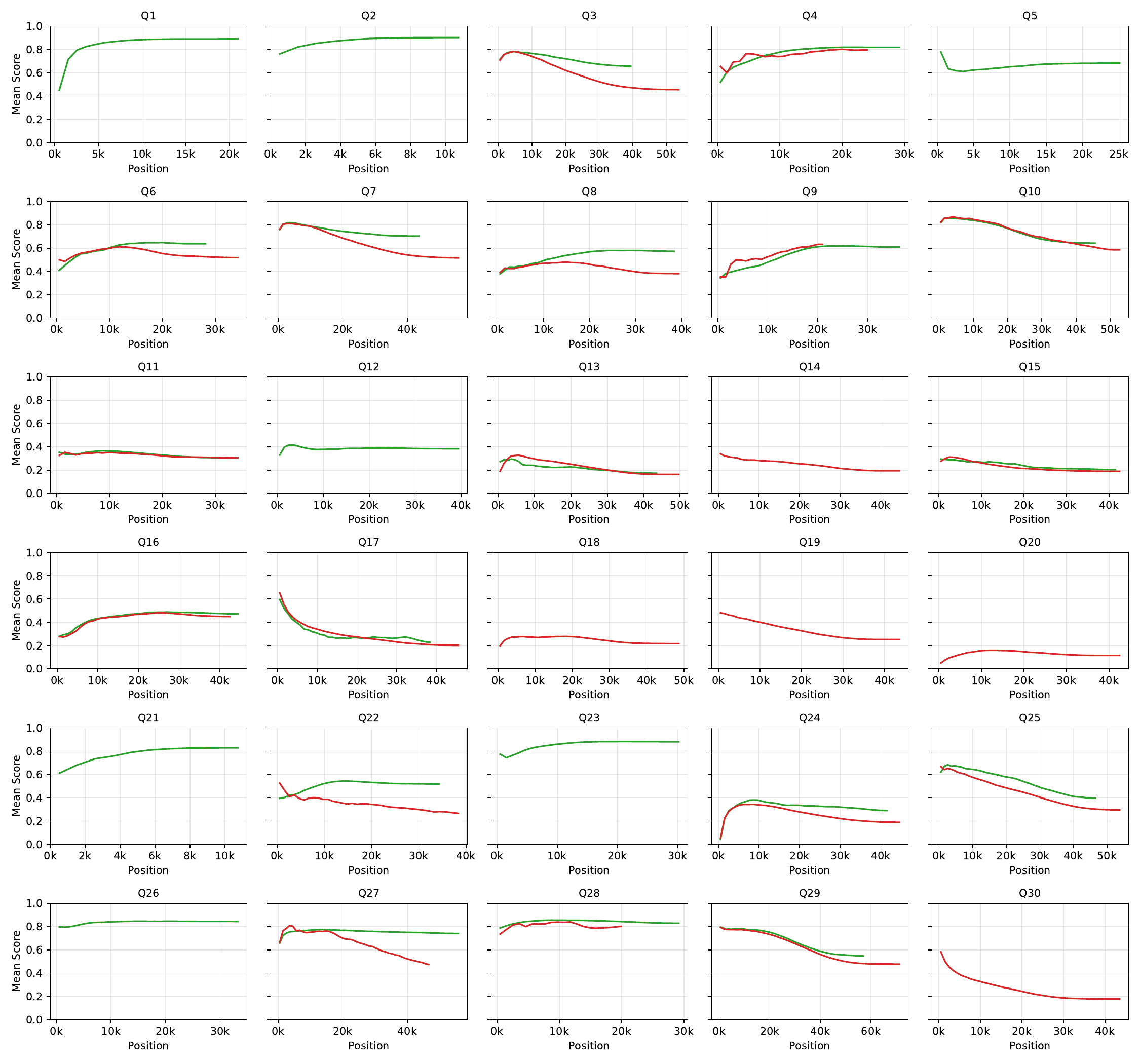}
    \caption{Trace-level score dynamics on AIME-25 for Qwen3-4B-Thinking-2507.}
    \label{fig:vis_4b}
\end{figure*}

\begin{figure*}
    \centering
    \includegraphics[width=1\linewidth]{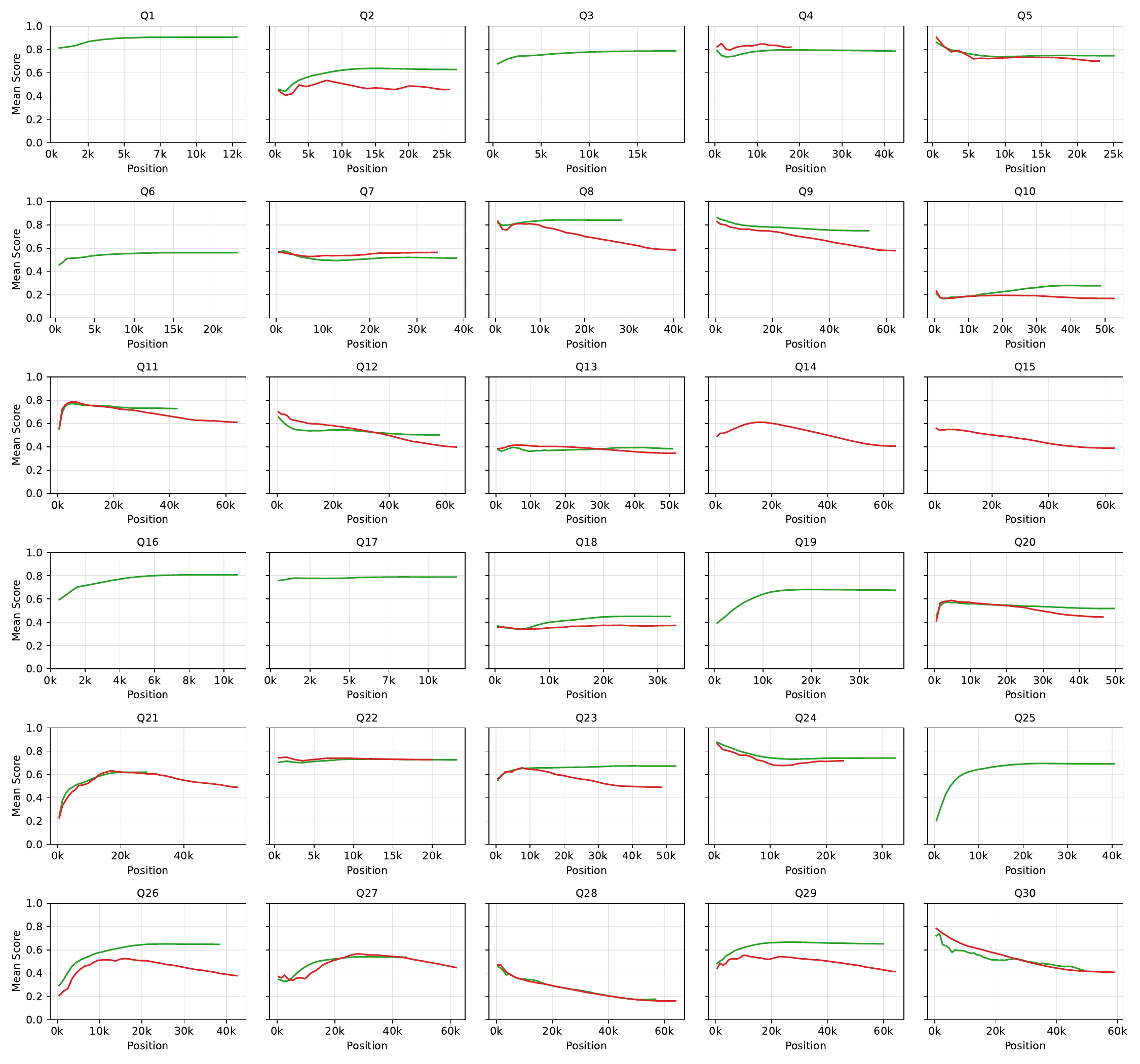}
    \caption{Trace-level score dynamics on AIME-25 for DeepSeek-R1-0528-Qwen3-8B.}
    \label{fig:vis_8b}
\end{figure*}

We visualize trace-level score dynamics on AIME-25 for Qwen3-4B-Thinking-2507 and DeepSeek-R1-0528-Qwen3-8B in Fig.~\ref{fig:vis_4b} and Fig.~\ref{fig:vis_8b}. Each subplot shows the prefix mean of step scores as a function of token position (grouped into 1024-token bins), where the green and red lines represent the average scores across correct and incorrect traces, respectively. Solutions are generated with N=64 samples per problem. The results demonstrate that our step scorer effectively separates promising reasoning paths from unpromising ones throughout generation.

\clearpage
\onecolumn

\setcounter{section}{0}
\renewcommand{\thesection}{\arabic{section}}

\end{document}